# Generating Information Extraction Patterns from Overlapping and Variable Length Annotations using Sequence Alignment

**Frank Meng**[1,2] **Craig A. Morioka**[3] **Danne C. Elbers**[1,4]
fmeng@bu.edu, danne.elbers@va.gov, cmorioka@ucla.edu
[1]Boston CSP Informatics, VA Boston Healthcare System
[2]Department of Medicine, Boston University School of Medicine
[3]Medical Imaging and Informatics Group, Department of Radiological Sciences, UCLA
[4]Complex Systems Center, University of Vermont

## Abstract

Sequence alignments are used to capture patterns composed of elements representing multiple conceptual levels through the alignment of sequences that contain overlapping and variable length annotations. The alignments also determine the proper context window of words and phrases that most directly impact the meaning of a given target within a sentence, eliminating the need to predefine a fixed context window of words surrounding the targets. We evaluated the system using the CoNLL-2003 named entity recognition (NER) task.

## 1 Introduction

We present a methodology that leverages a sequence alignment algorithm to automatically generate patterns consisting of heterogeneous annotations that can be applied to various natural language processing (NLP) problems such as information extraction (IE). The patterns generated integrate annotations that represent multiple types of semantics such as raw tokens, syntactic representations, and mappings from ontologies. Our pattern generation approach is capable of incorporating the information represented by various annotation types regardless of whether they are overlapping or exhibit variations in length. The central technique used is a modified Smith-Waterman approximate local alignment algorithm extended to align two-dimensional grids of annotations instead of one-dimensional sequences. For IE applications, we found that the sequences of words and phrases that are common between sentences that contain the same type of target are a good representation of the surrounding context likely to influence the meaning of the target. As validation of this work, we show that our approach performs comparably to established named-entity recognition (NER) models for the CoNLL-2003 Named Entity Recognition task.

## 2 Related Work

Information extraction has a rich history within the NLP community with numerous algorithms, methods, and systems developed to solve various aspects of the problem. Much attention has been paid to information extraction tasks in the form of conferences and competitions from domains ranging from military surveillance to biomedical informatics (e.g., Message Understanding Conferences (MUC) (Grishman and Sundheim, 1996), Text Analysis Conferences (TAC) (Ellis et al., 2016), and Biocreative (Krallinger et al., 2017)). Earlier work in IE began with techniques based on manually or semi-automatically generated extraction patterns (e.g., Appelt et al., 1995; Riloff, 1996; Soderland, 1999; Califf and Mooney, 1999; Brauer et al., 2011). More recently, there have been some further development of purely pattern-based systems as well as hybrid systems that combine patterns with machine learning, especially within the biomedical informatics domain (e.g., Gooch and Nguyen et al., 2010; Roudsari, 2012; Gupta et al., 2014; Tahsin et al., 2016; Meystre et al., 2017). However, most current IE work has focused on statistical modeling methods, such as Support Vector Machines and Conditional Random Fields, and are typically applied to specific tasks such as named entity recognition (e.g., Ju et al., 2011; Leaman et al., 2015; He and Grishman, 2015; Hassanpour and Langlotz, 2016; Kang et al., 2017; Gao et al., 2017). In addition, much recent effort has also been focused on applying neural methods to IE (e.g., Narasimhan et al., 2016; Limsopatham and Collier, 2016; Wei et al., 2016; Li et al., 2017; Palm et al., 2017).

Though statistical machine learning and neural methods have proven to be very effective, pattern or rule-based IE systems possess some specific advantages such as being easier to understand and more straightforward to maintain by both experts and non-experts alike. Furthermore, these systems are also more likely to exhibit higher precision since patterns capture specific word sequences that map to particular meanings but may cover fewer cases per pattern. Chiticariu et al. (2013) showed that the vast majority of IE systems deployed within large industrial software systems are pattern-based mainly because of these advantages, whereas most academic research systems are implemented using statistical machine learning. In particular, systems that can be easily interpreted by non-experts are important for domains such as medicine, where undetected system errors have the potential of resulting in serious consequences and possibly loss of human life. Thus, there is still a definite need to continue developing and refining pattern-based NLP methods and researching more powerful methods for synergistically integrating them with other machine learning techniques.

As will be discussed in more detail in later sections, the techniques developed in this work address issues that are important for increasing the utility of pattern-based IE systems. First, extraction patterns are automatically generated from the same annotated data sets used by typical machine learning algorithms and no additional manual crafting or curation is required in general. Second, though many pattern-based systems can generate high precision patterns based on sequences of tokens taken directly from the original text, they often lack a straightforward mechanism for generalizing these patterns with more abstract concepts. Our methodology generates patterns that represents multiple conceptual levels within the same sequence, from raw tokens to higher level semantic types, enabling a mechanism for pattern generalization that is built into the generation process. The methodology integrates overlapping and conflicting annotations provided by various sources, such as mappings from ontologies or direct manual labeling by experts. This closely follows certain usage-based linguistic constructs, such as pivot schemas, that are utilized by humans for generalization during language acquisition as multiple examples of similar linguistic constructs are encountered over time (Tomasello, 2005). Third, many existing pattern-based and machine learning IE algorithms leverage a fixed window of tokens surrounding the extraction target to identify contextual features. Similar to Meng and Morioka (2015), the sequence alignments utilized by our pattern-based system provide a natural mechanism for identifying the surrounding words and phrases that directly impact the meaning of the target without the need for a predetermined fixed window size. This is a more accurate model of linguistic structure, since the size of context windows is variable in general and depends on syntactic and semantic structure. Finally, pattern-based systems have the advantage over other machine learning techniques in that features do not need to be specifically selected and engineered for each task being performed. The patterns themselves leverage the inherent characteristics of the language based on the available annotations to determine the salient features that help to identify the extraction targets. The system we present in this paper could be applied to other IE and NLP tasks without needing to design and build another classifier from the ground up. The only processing done specifically for NER that we incorporated into our system was based on the rule of thumb that nouns were very likely to be classified as person names if other instances of the same noun within the same document were also classified as a person name. Other existing NER systems readily take advantage of this rule, so this gave our system no special or unfair advantage.

## 3  Methods

This section describes our definitions for overlapping and variable length annotations, extensions to the Smith-Waterman algorithm for handling these annotations, and methods for generating, filtering, and applying patterns.

## 3.1 Annotations

The most basic and commonly used device for augmenting the semantics of raw text is to layer annotations upon words or phrases. In this section, we give a brief working definition of how annotations are used and what assumptions were made in the work presented in this paper. Annotations are generated either computationally or manually and are associated with some error rate regardless of the source. More specifically, we define an annotation to be a tuple of the form: {*document_id*, *start*, *end*, *type*, *features*}, where the *document_id* uniquely identifies the document that the annotation is associated with, *start* is the start index, *end* is the ending index, *type* represents meaning of the annotation, and *features* capture other information specific to the annotation type. For instance, a *Token* annotation type may have additional features such as the root form or part-of-speech. Generally, annotations cover one or more words and can be overlapping, even among annotations of the same type, due to having multiple legitimate categorizations for the same words or phrases, ambiguities in the language, or annotation errors.

## 3.2 Alignments of Overlapping and Variable-length Annotation Sequences

Most typical alignment algorithms are designed to operate on sequences that consist of non-overlapping elements of fixed length (e.g., a string of characters or a sequence of words). Our extension to the Smith-Waterman approximate local alignment algorithm (Smith and Waterman, 1981) when used for processing text data with annotations relaxes these restrictions and is capable of aligning sequences of elements that are both overlapping as well as variable in length. In the case of overlapping elements, each cell within the alignment matrix will now represent multiple annotations and these must all be taken into consideration when determining matches. Since sequence elements are not all of the same length and may span multiple cells in the alignment matrix, each cell must now keep track of multiple scores as well as links that indicate multi-cell matches. An example of overlapping and multiple length annotations for a simple sentence is shown in Figure 1.

Lengths of elements are measured with respect to an annotation type designated to be atomic (e.g., the first row of words in Figure 1), where they cannot be further broken down into smaller units. The *Token* annotation is typically considered atomic because other annotation types often consist of one or more tokens. The atomic annotations themselves are defined to be of length one and elements that consist of multiple atomic annotations will be assigned lengths that correspond to the number of atoms that are included.

| The | man | sued | Acme | Company |
|---|---|---|---|---|
| :det | :noun | :verb | :nnp | :nnp |
| :noun phrase | | | :noun phrase | |
| | | :verb phrase | | |
| | | | :organization | |

Figure 1. Example of overlapping and variable length annotations for a sentence. These annotations may complement or possibly conflict with each other.

To describe the algorithm in more detail, we begin with the original Smith-Waterman definition of the alignment matrix for two sequences indexed by *i* and *j*:

$$M[i][j] = \max \begin{cases} 0 \\ M[i-1][j-1] + S[s_i][t_j] \\ M[i-1][j] - d \\ M[i][j-1] - d \end{cases}$$

Briefly, the matrix values are determined by generating scores for cells based on scores from previous cells, starting from the upper left corner, where higher scores represent matches between elements of the two sequences. Links are associated with each cell to enable backtracking through the matrix for generating the final alignment sequence. Gaps can be inserted to stretch the alignment in order to produce a best fit. An example matrix with scores and links for two simple sequences is shown in Figure 2 (diagonal arrows represent matches and verticals or horizontals represent gaps).

|   | A | B | C | D | E |
|---|---|---|---|---|---|
| H | 0 | 0 | 0 | 0 | 0 |
| A | 1 | 1 | 1 | 1 | 1 |
| B | 0 | 2 | 2 | 2 | 2 |
| G | 0 | 2 | 2 | 2 | 2 |
| C | 0 | 2 | 3 | 3 | 3 |
| D | 0 | 2 | 3 | 4 | 4 |

Figure 2. Smith-Waterman alignment matrix showing backtracking.

Handling overlapping and variable-length elements requires additional bookkeeping as the alignment matrix is being generated. Instead of aligning two sequences of annotations, the algorithm aligns two two-dimensional *grids* of annotations. Each cell in the matrix now contains a numeric score, backtracking link, and an additional *span*. Spans are the mechanism used for handling variable length elements and enable the backtracking phase of the algorithm to jump over matrix cells when elements cover more than one cell. Spans are visualized as links that join two cells and indicate the rectangle of cells in the matrix that are covered by any pair of elements $(x, y)$, where $x \in X$ and $y \in Y$. More specifically, the span representing the element pair $(x_i, y_j)$, where $x_i$ starts at position $i$ in $X$ and $y_j$ starts at position $j$ in $Y$, originates from the cell at $(i + len(x_i), j + len(y_j))$ and terminates at the cell $(i, j)$, where the function $len(e)$ returns the length of the element $e$ in terms of the number of atomic elements (or cells in the matrix). The rectangle of cells represented by this span has four corners at positions $(i, j)$, $(i + len(x_i), j)$, $(i, j + len(y_j))$, and $(i + len(x_i), j + len(y_j))$. The numeric score is stored at the span's origin and the backtracking link is located at the span's terminal.

For handling overlapping elements, instead of matching only one pair of elements at the current position $(i, j)$, the algorithm must now pairwise match the set of all elements of grid $X$ that begin at position $i$ ($x_i$) with all elements of grid $Y$ that begin at position $j$ ($y_j$). Each matching pair of elements $(x_i, y_j)$ will generate a score, backtrack link, and span. The score is stored at cell $(i + len(e_i), j + len(e_j))$, the backtrack link is stored at cell $(i, j)$, and the span, as described above, will originate from $(i + len(e_i), j + len(e_j))$ and terminate at $(i, j)$.

Figure *3* illustrates the pairwise matching process between two grids when determining the score for the current cell (upper left, bolded), where grid X is horizontal and Y is vertical. Multiple scores are generated for the current cell resulting from matching multiple pairs of elements that begin at the current cell (e.g., (A, A), (J, A), (B, A), (C, G) etc.). The highest numeric score will be kept for all spans that share the same origin and terminal. Notice that the score for element pair (B, A) is stored at cell (1, 0) because the element B has length 2. Figure 4 shows the score, backtrack link and several spans that can be generated from the same cell (upper left, bolded), with all spans sharing the same terminal cell but having different origin cells that result from other elements that also begin at the same cell, such as *B* and *C*.

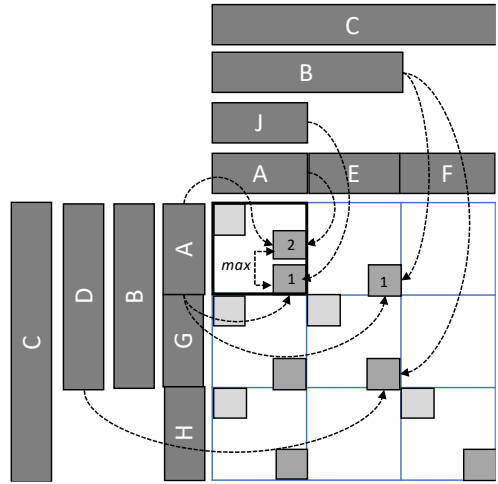

Figure 3. Pairwise matching between overlapping elements (e.g., annotations). The algorithm handles multiple possible matches for each cell for all elements that begin at that cell (top left).

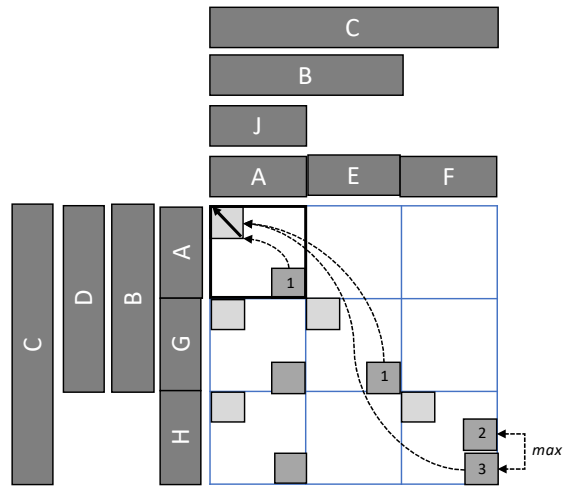

Figure 4. Multiple spans associated with the top left cell (bolded). Scores are stored at the source of the span and backtrack links are stored at the terminal of the spans. The span utilized is associated with the current maximum score.

As with the original alignment matrix definition, backtracking links are generated based on which adjacent cell contributes to the highest score for the current cell. As mentioned previously, the backtracking link is stored at the termination point of each span. The overall backtracking process in the extended algorithm begins at the cell holding the globally maximum score. For each cell, the traversal process begins at the origin of the cell's span, follows the span's link to its terminal cell where a backtracking link is located, and then follows the backtracking link to the next cell that points to the origination of that cell's span. This process is repeated until a score of 0 is encountered. Instead of stepping through the matrix cell by cell as with traditional alignment algorithms, the modified algorithm traverses from one span to the next. During backtracking, co-occurrence of elements is captured by identifying matching elements with the same start and end positions. Element co-occurrences can model cases where the correlation of multiple elements together is a better representation of the language pattern. A simple example of this would be a word and its part-of-speech, where the same word can be used for different parts-of-speech and correlating a word with its part-of-speech provides more information. The described process is illustrated in Figure 5, where the function of spans is seen in guiding the algorithm to jump over multiple matrix cells based on the lengths of matched sequence elements. The backtracking process also captures the co-occurrence of elements *B* and *E* and expresses this information in the resulting alignment shown at the bottom of the figure.

### 3.3    Using Local Sequence Alignment to Generate Extraction Patterns

Given a document corpus *D* and a set of annotations for the documents in *D*, $A_D$, $T \subseteq A_D$ is defined to be the set of *target* annotations, a specific annotation type reserved to represent the targets of the IE task at hand. Annotations in *T* represent words or phrases of interest that are being extracted. For every instance $t \in T$, the set of annotations in its surrounding general context, $GC_t \subseteq A_D$, can be defined as the window of words surrounding the position of *t* in a section of text that will be used for generating a specific context $C_t \subseteq GC_t$ which is a candidate minimal window of words that exert an influence on the meaning of *t* with respect to the extraction task at hand.

In practice, we have found that using the entire sentence containing the instance *t* as the $GC_t$ is a reasonable approach, but utilizing a window of tokens of arbitrary size that precedes and follows each instance (possibly spanning multiple sentences) is another possible choice. Pairwise alignment is performed between $GC_t$'s for all $t \in T$ and results in a corresponding $C_t$ for each *t*, where $C_t$ could be the empty set {}. For the purposes of this work, the scoring matrix used by the alignment algorithm is tuned to favor alignments that include instances of *T* by setting the match score for *T* to be much higher than matches for other annotation types. Setting the *T* match score to be 100, other matches to be 1, mismatches to be -1 and a high gap penalty allowed us to generate useful patterns. Besides guaranteeing the inclusion of each target, this configuration rewards all matches and favors gaps over mismatches. Figure 6 shows several sentences containing instances of targets with their respective varying *GC*'s and the results of local alignment. The figure illustrates that the alignment process identifies the same $C_t$ for each target *t* even though they all have differing $GC_t$'s, thus greatly reducing the level of lexical variation and complexity and potentially narrowing down the window of surrounding words that indicate the meaning of *t*.

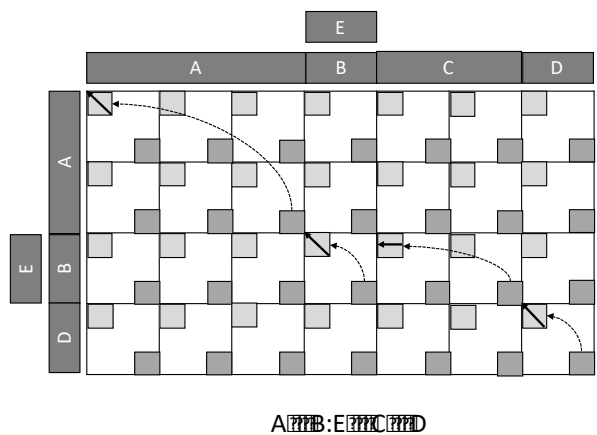

Figure 5. Backtracking in the alignment matrix using backtrack links and spans, illustrating the use of spans to jump over multiple cells based on the lengths of elements.

### 3.4    Target Patterns

The words and phrases that constitute targets can also be described by patterns, enabling the system to capture generalizations of the information being extracted. Certain types of targets, such as social security numbers or zip codes, can be rigorously specified by formal methods such as regular expressions, making them trivial to identify within text. Generally speaking, however, targets are composed of complex multi-word phrases that do not follow simple lexical patterns and structures. More specifically, each grid of annotations that is covered by a given $t \in T$ is pairwise aligned with all other target grids using the same extended Smith-Waterman algorithm resulting in a set of patterns that characterize targets. This alignment process differs slightly from the one described previously for context patterns because they clearly do not include a target annotation. Otherwise, the alignment mechanism remains the same.

Two sets of patterns are generated by the processes described in this section: $P_C$, the set of all patterns that are based on the specific contexts $C$; and $P_T$, the set of all patterns that characterize the set of targets $T$. All $p_c \in P_C$ have three components: left context ($lc$), target, and right context ($rc$), where at most one context can be null. All $p_t \in P_T$ are grids of annotations that have no special structure.

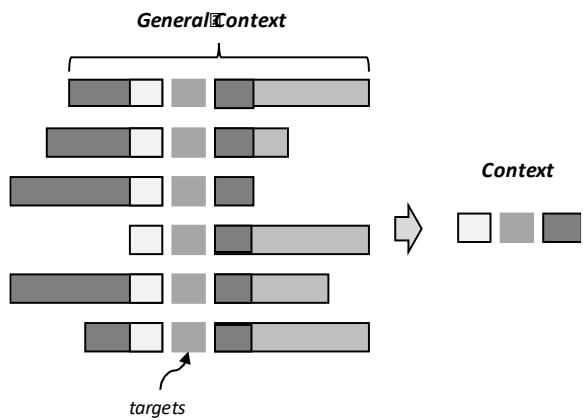

Figure 6. Generating extraction patterns using local alignment. As targets are aligned from various sentences with different General Contexts, commonalities are manifested and these are expressed as the specific Context for the targets (right).

## 3.5 Applying Patterns to Text

To extract targets from text using the set of patterns $P_C$, if $lc$ and $rc$ exist in $p_c \in P_C$, both are aligned against the text with the requirement that $rc$ must appear after the end of $lc$ in sequential order of tokens. If both contexts produce a valid alignment in the text, the tokens that fall after the end of $lc(p_c)$ and before the beginning of $rc(p_c)$ are considered to be a candidate target (Figure 7, top). If only an $lc$ exists, all text following the end of $lc(p_c)$ is considered as the candidate target, usually limited by the size of the general context $GC$ (Figure 7, bottom). Patterns with only an $rc$ are handled analogously. The candidate target is then aligned with all $p_t \in P_T$ and any successful alignment indicates the pattern has successfully identified a valid target.

In general, there may be interdependencies between patterns such that some patterns may depend on annotations generated by other patterns. For instance, certain patterns that extract last names of people may depend on other patterns that generate annotations indicating first names. Thus, these last name patterns will fail to match any sequences of annotations until the "first name" patterns have been able to generate their annotations. To handle cases such as these, the system will iterate through the exhaustive matching process over all types of extraction patterns, generating new annotations and terminating when no more new matches have been detected.

Pattern-target pairs are treated as a unit when being applied because the semantics of a pattern is dependent on the meaning of the target. For example, the pattern "visiting :target" (where the symbol *:target* represents any word or phrase that is being extracted as the target) can be a high precision pattern for extracting a destination if the target has been annotated as a geographical location, such as a country or city name (e.g., "visiting Boston"). If the target is simply a noun or another semantic category, the pattern's quality may not be high enough to retain. For instance, if the pattern is applied to the phrase

"visiting parents", it would erroneously extract "parents" as a destination location even though the pattern has perfectly matched the text. We define $PT_A$ as the set of all pattern-target pairs $(p, t)$, where $p \in P_C$ and $t \in P_T$, that extracts targets and generates annotations of type $A$.

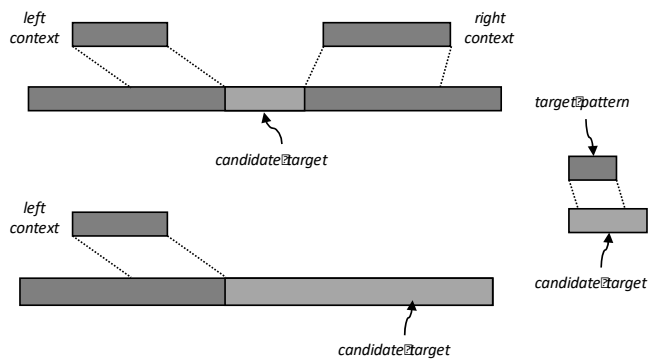

Figure 7. Applying patterns to sequences of text. Patterns generally have a left and right context with the candidate target falling in between (top). In some cases, patterns only have either a left or right context (bottom). The candidate target is matched against target patterns (right).

### 3.6 Capturing and Filtering Effective Patterns

The patterns that result from commonalities between sequences of text captured by the alignments will vary in generality and scope depending on the semantic similarity between the text. For instance, the commonalities between sentences that have very different semantic meanings will tend to represent syntactic structure and shorter phrases that will result in less specific patterns. The opposite is true for sentences that have very similar meanings, where more specific and longer phrases result from the commonalities between them. The pattern-target pair set $PT$ must be refined to retain only patterns that meet a minimum quality level determined by a goodness metric. In practice, we have found that a goodness metric that maximizes precision (PPV) works well in many applications because the recall (sensitivity) of the set $PT$ generally increases as more patterns are generated through additional training data. Also, high precision patterns enable the manipulation of intermediate results and other meta-analyses applied to the patterns themselves that can increase the performance of the system. More formally, the refinement process for $PT$ determines a score for each pattern-target pair $(p, t)$ that was applied on a set of training examples based on a chosen goodness metric, $M(p, t)$. In our case, the refined pattern set $PT' \subseteq PT$ is composed of all pairs that exhibit a precision above a user-defined threshold (typically 0.95 or above) were considered high quality.

Patterns often overlap one another and need to be filtered to obtain a minimal set of patterns that retains the same extraction performance of the original set. Filtering is performed using a simple heuristic where a pattern-target pair $(p, t)$ is filtered out of $PT'$ if every target extracted by $(p, t)$ is also extracted by a set of pairs $(p', t')$, where for every $p'$, $\text{len}(p') < \text{len}(p)$ and $(p', t') \neq (p, t)$.

### 3.7 Prior Knowledge

Incorporating prior knowledge to machine learning systems enables bootstrapping performance without the need to rely solely on knowledge garnered from training data. Our system integrates prior knowledge through annotations as well as prior probabilities calculated from training data. First, because the system can incorporate any annotation regardless of whether it overlaps or conflicts with other annotations, the addition of new annotation types adds knowledge to the system. The NER application used for evaluation (See Section 4) utilized the GATE NLP framework to automatically generate annotations that were used for the extraction patterns. Table 1 shows the GATE annotation types used for NER. For the Token type, the root feature provides the root form of the token, string is the token's raw string, and category is the part-of-speech.

Second, the system utilizes prior probabilities estimated from the training data. Prior probabilities were effective for this particular task because names of persons, organizations, and locations have high semantic content in themselves and often can be identified without support from the surrounding context. More specifically, the system utilizes prior estimates to both apply new labels to tokens with high priors as well as remove labels with low priors that were automatically annotated by the system.

| Annotation Type | Feature(s) |
|---|---|
| Token | root, string, category |
| Date | normalized |
| Number | value |
| Lookup | majorType |

Table 1: GATE annotation types

## 4 Evaluation

We evaluated our pattern-based system for labeling proper nouns that correspond to names of persons, organizations, and locations in the 2003 Conference on Computational Natural Language Learning shared task (CoNLL 2003) training corpus (Sang et. al., 2003). Following standard practice, we generated, refined, and filtered our pattern-target pairs using documents in both the provided training data and test A corpus, and then evaluated the resulting pattern-target pairs on documents in the test B corpus. In addition, precision and recall scores were all measured at the entity level, where entities that were labeled with extra tokens or found to have missing tokens were counted as both false positive and false negative.

Table 2 shows the performance of our system compared with the Stanford CRF NER system (Finkel et. al., 2005). In particular, we found that the F1 scores for the three categories of person names (*PER F1*), organizations (*ORG F1*), and locations (*LOC F1*) were comparable to Stanford's reported performance (see row labeled *Entity*). The *Token* row shows the precision and recall scores for our system calculated at the token level. The *Patterns, Token* row shows the results of only applying the patterns without the use of prior knowledge. Finally, the *Lookup* row shows the performance of GATE's Gazetteer when applied to the same extraction task using only those annotations that are related to person names, organizations and geographic locations.

|  | PER Prec | PER Recall | PER F1 | ORG Prec | ORG Recall | ORG F1 | LOC Prec | LOC Recall | LOC F1 |
|---|---|---|---|---|---|---|---|---|---|
| Stanford |  |  | 0.923 |  |  | 0.817 |  |  | 0.885 |
| Entity | 0.933 | 0.895 | 0.914 | 0.88 | 0.736 | 0.802 | 0.886 | 0.858 | 0.872 |
| Token | 0.961 | 0.913 |  | 0.905 | 0.758 |  | 0.845 | 0.871 |  |
| Patterns, Token | 0.97 | 0.661 |  | 0.927 | 0.481 |  | 0.904 | 0.653 |  |
| Lookup | 0.914 | 0.35 |  | 0.74 | 0.142 |  | 0.282 | 0.808 |  |

Table 2: Results for CoNLL-2003 NER Extraction Data Set

| Pattern (PER) | Target | Count |
|---|---|---|
| :target :token\|category\|nnp :token\|string\|(!:token\|root\|(!:token\|category\|( :lookup\|majortype\|location | :lookup\|majortype\|person_first | 124 |
| :token\|category\|nnp!:i-per :target :token\|string\|(!:token\|root\|(!:token\|category\|( | :token\|category\|nnp | 94 |
| :token\|string\|,!:token\|root\|,!:token\|category\|, :lookup\|majortype\|person_first :target | :token\|category\|nnp | 63 |
| :target :token\|category\|nnp!:i-per :token\|string\|(!:token\|root\|(!:token\|category\|( | :token\|category\|nnp | 52 |
| :lookup\|majortype\|person_first!:i-per :target | :token\|category\|nnp | 36 |
| :i-per :target :token\|string\|,!:token\|root\|,!:token\|category\|, | :token\|category\|nnp | 35 |
| :lookup\|majortype\|jobtitle :target | :token\|category\|nnp!:lookup\|majortype\|person_first | 34 |
| :token\|category\|nnp :target :token\|string\|(!:token\|root\|(!:token\|category\|( :token\|string\|germany!:token\|root\|germany!:token\|category\|nnp!:lookup\|majortype\|location :token\|string\|)!:token\|root\|)!:token\|category\|) | :token\|category\|nnp | 34 |
| :target :token\|category\|nnp!:i-per | :token\|category\|nnp!:lookup\|majortype\|person_first | 32 |
| :token\|string\|-!:token\|root\|-!:token\|category\|: :target",":token\|category\|nnp :token\|string\|,!:token\|root\|,!:token\|category\|, | :lookup\|majortype\|person_first | 28 |

Table 3: Top frequently used patterns for person names

| Pattern (ORG) | Target | Count |
|---|---|---|
| :start :target :number :number :token\|category\|cd!:number | :token\|category\|nnp!:lookup\|majortype\|location | 146 |
| :start :target :token\|category\|nnp :number :number :number | :token\|category\|nnp!:lookup\|majortype\|location | 42 |
| :target :token\|string\|at!:token\|root\|at!:token\|category\|nnp :lookup\|majortype\|location :end | :token\|category\|nnp | 39 |
| :lookup\|majortype\|organization :target :lookup\|majortype\|organization!:token\|category\|nnp | :lookup\|majortype\|organization | 31 |
| :start :token\|category\|nnp :target :number :number :number | :lookup\|majortype\|location | 24 |
| :start :target :token\|category\|nnp!:lookup\|majortype\|location :token\|category\|cd!:number :number | :token\|category\|nnp | 24 |
| :start :target :token\|category\|nnp!:lookup\|majortype\|location :token\|category\|cd!:number :number | :token\|category\|nnp!:lookup\|majortype\|location | 23 |
| :target :token\|category\|cd!:number :token\|category\|nnp!:lookup\|majortype\|location!:i-org | :token\|category\|nnp | 23 |
| :start :token\|category\|nnp!:lookup\|majortype\|location :target :number :number | :token\|category\|nnp!:lookup\|majortype\|location | 20 |
| :target :token\|category\|nnp :number :token\|category\|cd!:number :number :number :token\|category\|cd!:number | :token\|category\|nnp | 20 |

Table 4: Top frequently used patterns for organization names

| Pattern (LOC) | Target | Count |
|---|---|---|
| :start :target :number :token\|string\|-!:token\|root\|-!:token\|category\|: :number :end | :token\|category\|nnp | 143 |
| :target :lookup\|majortype\|currency_unit | :token\|category\|nnp!:lookup\|majortype\|location | 62 |
| :target :number :token\|string\|-!:token\|root\|-!:token\|category\|: :number :end | :token\|category\|nnp!:lookup\|majortype\|location | 56 |
| :target  :token\|string\|]!:token\|root\|]!:token\|category\|] | :token\|string\|germany!:token\|root\|germany!:token\|category\|nnp!:lookup\|majortype\|location | 35 |
| :lookup\|majortype\|person_first  :token\|category\|nnp  :token\|string\|(!:token\|root\|(!:token\|category\|(  :target  :token\|string\|)!:token\|root\|)!:token\|category\|) | :lookup\|majortype\|location | 29 |
| :token\|string\|(!:token\|root\|(!:token\|category\|(  :target  :lookup\|majortype\|location | :token\|category\|nnp!:lookup\|majortype\|location | 25 |
| :lookup\|majortype\|currency_unit!:i-loc  :target | :lookup\|majortype\|currency_unit | 25 |
| :target  :token\|string\|]!:token\|root\|]!:token\|category\|] | :token\|string\|france!:token\|root\|france!:token\|category\|nnp!:lookup\|majortype\|location | 25 |
| :lookup\|majortype\|person_first   :token\|string\|(!:token\|root\|(!:token\|category\|(  :target | :token\|category\|nnp!:lookup\|majortype\|location | 25 |
| :start  :token\|category\|nnp  :token\|string\|at!:token\|root\|at  :target  :end | :lookup\|majortype\|location | 24 |

Table 5: Top frequently used patterns for location names

Tables 3-5 show the ten highest frequently used patterns for each label type, where the first two columns show the pattern-target pairs. The patterns are shown as sequences of elements, where each element consists of multiple sub-elements that are divided by the exclamation point symbol (*!*). Each sub-element, in turn, is composed of a series of labels demarcated by the pipe symbol (*/*) that indicate the annotation type, the feature of the annotation (if applicable), and the value of the feature (if applicable). For instance, the top pattern for person names shows the target occupying the first position in the sequence. The second element *:token/category/nnp* consists of one sub-element consisting of three labels that indicate the annotation type is a token (*:token*) with the feature *category* that has the value *nnp* (indicating a proper noun). The third element *:token/string/(!:token/root/(!:token/category/(* is more complex, composed of three sub-elements, where the first element is of type *:token* with feature *string*, having the value of an open parenthesis ("("). The second sub-element represents the *root* feature of the *:token* annotation type with value open parenthesis. The *root* feature is the root representation of a word (e.g., heal vs. healed). Since this particular example shows an open parenthesis symbol, there is no difference between the actual string representation and the root. The third sub-element indicates the *category* feature of the *:token* annotation type for this token. Sub-elements indicate co-occurrences between two tokens at the same position of the alignment as described in Section 3.2. An example of an element with only one sub-element and one label is shown in the top pattern for organization names, where the element *:number* indicates the annotation type *:number* (representing a numeric value) having no associated labels indicating features. As mentioned in a previous section, the target pattern associated with a pattern also plays a major role in the performance of that pattern. For the top person name pattern, the associated target pattern *:lookup/majortype/person_first* indicates a person's first name, where the annotation type is *:lookup* (GATE's Gazetteer) with the *majortype* feature that has the value *person_first*. The *:target* element is a special symbol indicating the position of the target in the pattern sequence, *:start* indicates the start of a sentence, and *:end* indicates the end of a sentence.

## 5 Discussion

Our pattern-based system compared favorably with Stanford's NER system on the CoNLL-2003 data set based on the entity-level F1 scores for person, organization, and location names. Token-level performance metrics were also very similar to their entity-level counterparts. It is interesting to note the effect of prior knowledge on the system's overall performance, particularly recall levels. Pattern-only performance compared with patterns and prior knowledge indicate that using only patterns results in slightly higher precision at the token level, but prior knowledge adds a non-trivial amount of recall (row labeled *Patterns, Tokens*). This shows that a fair percentage of cases were identified solely using high probability cases from the test set. Low probability priors were also used as negative examples to help reduce false positive instances. This result is not surprising given the high semantic content of names that enables their identification without using much information from the surrounding context. Though GATE's Gazetteer provided many useful annotations, using only the Gazetteer's results would have produced low performance metrics for all three label types (row labeled *Lookup*). For person names, Gazetteer annotations exhibited high precision because of the previously mentioned high semantic content (particularly Western first names).

As can be seen, many of these patterns reflect the semi-structured nature of the news articles in the CoNLL-2003 corpus. For instance, the top frequently used pattern for person names typically matches articles that report results from international sporting events, where the athlete's name is followed by the country they are representing enclosed by parentheses. This reliance on semi-structured contexts, however, utilizes minimal semantic content and was shown to easily fail if the same structure is suddenly used to express something different. This specific pattern caused multiple false positives when applied to the test set because an article contained the same basic textual structure but instead of an individual person's name with their associated country, a sporting organization's name was used (e.g., Real Madrid (Spain)). Other patterns that depend heavily on inherent textual structure can be seen in the patterns for both organizations and locations. Several of the top organization patterns show the target embedded among several numeric values. These patterns match articles that report scores of sporting activities, particularly National Basketball Association (NBA) or National Football League (NFL) game results, where the sporting team's name is being extracted (e.g., Los Angeles Lakers). Several examples of second-tier patterns can also be seen that utilize results from annotations derived from previous pattern matches. The fourth top *person* pattern contains the element *:token/category/nnp!:i-per* that has a dependency on an existing person annotation (*:i-per*).

Currently, the annotation types included in the sequences are weighted according to expert assessment of importance. Since this weighting scheme is subjective in nature, we propose to utilize a more objective weighting mechanism in the future, such as generating relative weights based on the prevalence of the annotation type within the document set. Thus, very common annotation types (e.g., part of speech) would be weighted less than those that are less common (e.g., ontological categories).

The sequences being aligned were derived using sentences as a natural boundary to break up large blocks of text for the general contexts (GCs). However, for certain semi-structured news articles within the CoNLL 2003 training set, such as results of sporting events, the general context represented by sentence boundaries generated by the GATE system were not informative enough to provide differentiation between positives and negatives. To address these types of cases, the system would need to be equipped with the ability to assess a more global context of the training example in addition to the local context of the sentence (e.g., Patwardhan and Riloff, 2007).

We are planning to apply this methodology to generate patterns that are able to capture relations between concepts. Since most binary relations are indicated by the language that occurs between instances of concepts, the pattern generation process described in this paper should be adaptable to characterize this language given the location of the concepts of interest.

As mentioned previously, patterns fundamentally capture the dependencies between a concept of interest and the semantics of the words and phrases that surround it. Generally, semantically rich contexts produce higher quality patterns but inter-dependencies can occur when patterns that extract one concept depend on annotations produced by patterns that extract other concepts. We will explore an iterative approach to pattern generation, where the annotations provided by previous iterations would be incorporated by patterns in the current iteration.

Because patterns are relatively simple structures, it is feasible to perform analyses on the patterns themselves. We plan to develop frameworks for systematically generalizing patterns by automatically moving elements to higher levels of abstraction that can represent generalized concepts.

# 6 Conclusion

We presented an automatic pattern generation methodology that is capable of integrating information from multiple levels of semantics based on available annotations. Along with the other advantages of pattern-based systems including transparency and portability, our system has the capability to generate patterns that are able to model complex linguistic constructs given a rich enough annotation set with the required training data. We demonstrated that the system's performance for the CoNLL-2003 NER task is very comparable with a well-known existing NER system for extracting person, organization, and location names.